# Analysis on distribution and clustering of weight

Chunming Ye[1]    Wenquan Tian[1]    Yalan Gao[1]    Songzhou Li[1]
1 Suzhou University
{winshepherd@163.com, szxytianwq@126.com, gaoyalan1212@student.usm.my, songzhouli@ahszu.edu.cn}

**Abstract**

   The study on architecture and parameter characteristics remains the hot topic in the research of large language models. In this paper we concern with the characteristics of weight which are used to analyze the correlations and differences between models. Two kinds of vectors-standard deviation vector and clustering vector-are proposed to describe features of models. In the first case, the weights are assumed to follow normal distribution. The standard deviation values of projection matrices are normalized to form Standard-Deviation Vector, representing the distribution characteristics of models. In the second case, the singular values from each weight projection matrix are extracted and grouped by K-Means algorithm. The grouped data with the same type matrix are combined as Clustering Vector to represent the correlation characteristics of models' weights. The study reveals that these two vectors can effectively distinguish between different models and clearly show the similarities among models of the same family. Moreover, after conducting LoRA fine-tuning with different datasets and models, it is found that the distribution of weights represented by standard deviation vector is directly influenced by the dataset, but the correlations between different weights represented by clustering vector remain unaffected and maintain a high consistency with the pre-trained model.

## 1.Introduction

   Large language model (LLM) technology had been widely applied in scientific research, education, medical, art design and many other fields. Research on LLMs encompassed (Naveed et al., 2025) architectural, training strategies, context length, fine-tuning, multimodal LLMs, evaluation, datasets, benchmarking, efficiency, and more. In the relevant research about model performance, many papers focused on innovations in model architecture(Achiam et al.,2023)(Touvron et al.,2023)，improvements in training methods(Gu et al.,2024)(Kumar et al., 2025)，adjustments in parameters(Xiao et al., 2023)(Sun et al.,2023)(Black et al.,2022) , and so on. Some studies concentrated on the correlation between weight parameters and model performance. Early research analyzed the characteristics of the weights themselves. It was generally believed (Lee et al., 2017)(Lee et al., 2023)(Yang, G, 2019) that weights and parameters in many neural network models were equivalent to a Gaussian process or a mixture of Gaussian distributions in the limit of infinite network width. The weight matrices of trained deep neural networks were studied with random matrix theory (Thamm et al., 2022). The statistics indicated that these weights themselves were random and did not contain any specific information. And the spectral distribution of large singular values of weight matrices were found that they cannot be defined by a tail index.

   With the emergence of different LLMs, some research attempted to manipulate the weights of the models to adapt to various application scenarios. The most common method (Jin et al., 2024) was parameter quantization techniques, which reduce the bits needed for model weights or activations. Some methods were used to compress models such as lower-precision format (Banner et al., 2019)(Dettmers et al., 2021), quantize both activations and weights as 8-bit integers during training procedure(Jacob et al., 2018)(Park et al., 2018) , or design post-training quantization algorithms(Liu et al., 2021)(Shang et al., 2023) for reducing the memory storage and computational costs. Sparsification was also used to simplified the weights in models. The method which removed some less important weights and structures from models was called weight pruning (Frantar et al.,

2023)(Sun et al., 2023). The sparse attention mechanism(Zaheer et al., 2020)(Beltagy et al., 2020) was also used to reduce computational overhead while maintaining model performance. Weight initialization was another approach employed for training models. Random weight initialization (Cao et al., 2018) was widely employed in neural network models. Gaussian process (Lee et al., 2017), orthonormal Basis(Saxe et al., 2018), Bayesian approach(Murru et al., 2016) and other methods were introduced to initialize weights. It was also found that in fine-tuning (Dodge et al., 2020) weight initializations produced by different random seeds would lead to different training results.

With the in-depth analysis on weights, researchers had found that not all weights were equally important for LLM. Pruned only part of weights in mlp.down.proj matrices, Llama-7B model's ability to generate text would be completely destroyed. These outliers in LLM were called super weights(Yu et al., 2024). These few super weights were commonly present in many models, and they had a significant impact on model performance. The study(Lin et al., 2024) which reduced model resource consumption by LLM low-bit weight-only quantization found that protecting 1% of salient weights could greatly reduce quantization error. Other study (Sun et al., 2024) observed that in various LLMs very few activations exhibited significantly larger values than others. Experiments showed that setting these activation values to zero may lead to a significant decline in model's performance.

It was evident from the existing research that: current large language models shared similar architectures and worked on analogous principles. Operations about some weights such as pruning, sparsification, and initialization had significant effects on model training and inference. Experimental analysis also showed that although the weights in various models all exhibited Gaussian distribution, there were considerable differences in the characteristics of weights even within the same model, leading to different impacts on the model working mechanism. Uncovering these latent weight signatures would provide more insights for model optimization.

Our contributions are the following:

(1) Weights are first extracted from the Query, Key and other projection matrices of a model and assembled into separate groups. Inspection shows that distributions of weights in these groups differ in shape. Then a standard-deviation vector is constructed by computing and normalizing the standard deviation values of each group. This vector serves as characteristic of the model weight. Experiments reveal that the profiles of these vectors are clearly distinct across different model families, yet remain obvious similar within the same family. The standard-deviation vector can be used as one of practical fingerprints for model identification.

(2) We use K-Means algorithm to perform cluster analysis on the singular values of weights in the projection matrices of each layer in model. Grouping the clustering results according to types such as Query, Key etc. and using the grouped results to form clustering vectors for constructing weight characteristic of the model. The analysis shows that the clustering vectors of models from the same model family are almost identical, while the clustering vectors between different model families differ significantly. The clustering vector can be used as one of characteristics to distinguish LLMs.

(3) Using the standard-deviation vector and clustering vectors as dual lenses, we profile the A and B matrices produced by LoRA fine-tuning. The shape of standard-deviation vector is closely related to the training corpus. Different models fine-tuned on the same dataset converge to almost identical profiles, regardless of their original architectures. In contrast, the clustering vector faithfully inherits the pre-trained model's feature, regardless of their dataset. These two kinds of vectors offer reference ideas for analyzing the relationship between models and weights.

## 2. Distribution of Weights

Weight distribution is one of characteristics for pre-trained model, which may be related to the performance of the model. We extract weights from model in different layers and analyze the difference among these distribution characteristics. The analysis is concerned with Query, Key, Value, Output projection matrices for attention mechanism weights, and gate-projection, up-projection, down-projection matrices for feedforward network weights. The weights of the whole model have been explored in previous studies. We will skip the analysis of the overall weights and instead focus on examining the distribution of different types of weights and the relationships between them.

### 2.1 Distribution of different projections

We select the following models for study: LLaMA 3.2-1B/3B/8B-Instruct(Touvron et al.,2023), SmolLM2-135M/1.7B(Allal et al., 2025), and Qwen2.5-1.5B/7B(Bai et al., 2023). These models have similar architectures with Q, K, V and other projection matrices, we extract and analyze weights projection-by-projection, treating each projection matrix as the basic unit. Taking LLaMA 3.2-1B as an example, the model comprises 16 transformer layers. There are self-attention blocks with Q-proj, K-proj, V-proj, and O-proj and MLP blocks with gate-proj, up-proj, down-proj in each layer. Each projection matrix in the layer is composed of the tensors of weights. To construct Q-weight matrix for model, we extract weights from the Query-projection matrix of each layer, concatenate all 16 layers' Query weight matrices, and form the final Query-Matrix. The same steps are followed to construct the Key-Matrix, Value-Matrix, Output-Matrix, Gate-Matrix, Up-Matrix and Down-Matrix respectively. Figure 1(a) shows the probability distribution of the weight values in Q, K, V, and O Matrix of LLaMA3.2-1B. Figure 1(b) shows the probability distribution of the weight values in the gate, up, and down matrix. We randomly selected 2000 weight values from each matrix for plotting. Only selecting part of the data for plotting has a minimal impact on the graphics, but it does not affect our conclusions. The figure shows that the distributions of Q, K, V, and O weights differ markedly, whereas the gate, up, and down weight profiles are much more similar with each other. We conjecture that the distinct functional roles of the corresponding blocks in the model sculpt these characteristic patterns.

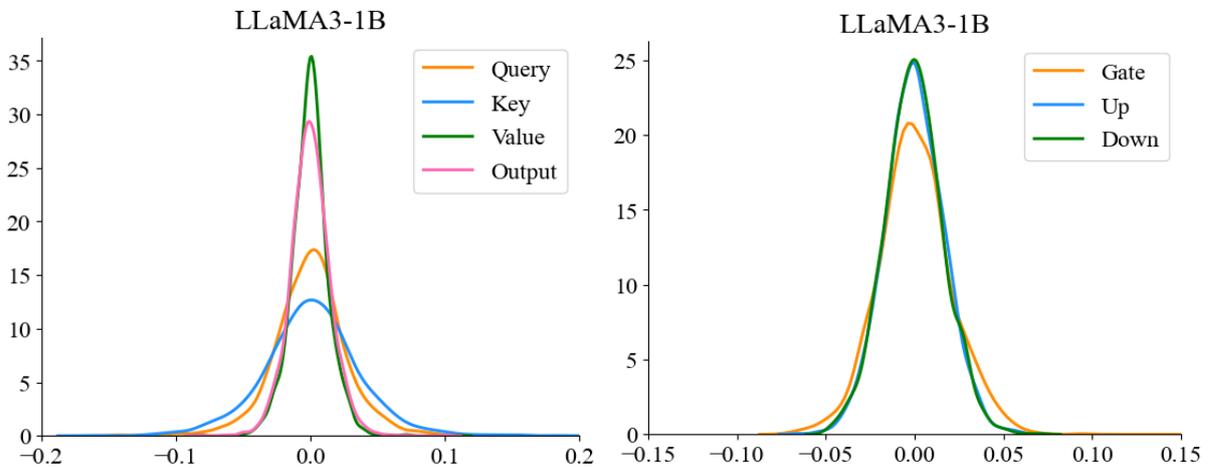

Figure 1 (a) Distribution of Q, K, V, O Matrix     (b) Distribution of Gate, Up, Down Matrix

The same analysis are performed on SmolLM2 and Qwen2 models. The results are shown in Appendix Figure 11. In these models the weight distribution have little differences in shape, but all of them conform to the expected Gaussian/Normal distribution. Then for the Query-Matrix data, comparisons are made between different models and between models in the same family but with different scales. Figure 2(a) shows the differences among the SmolLM2-1.7B, LLaMA3-1B, and Qwen2-1.5B for the distribution shapes of Query-Matrix. The weight values in Query-Matrix of the SmolLM2 model are distributed more uniformly, while the values in the LLaMA3 and Qwen2 models are more concentrated. Figure 2(b) shows that within the models in the same family, the shapes of the weight distributions are relatively similar. However, there are also differences in the degree of concentration of the values. Although only 2,000 data points are selected for plotting in each case, the weight values of the larger-scale LLaMA3-8B model are more concentrated. Similar results are observed from the analysis of other matrices including Key-Matrix, Down-Matrix and so on. Some distribution of matrices are plotting in the Appendix Figure 12.

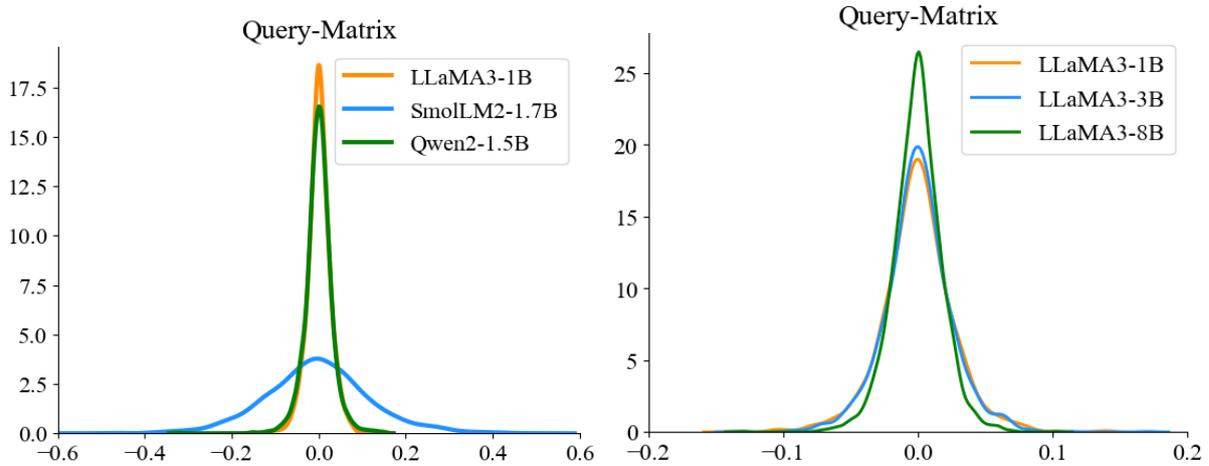

Figure2 (a) Distribution of different models   (b) Distribution of models with different scales

## 2.2 Standard-Deviation Vector

Analysis above shows that for each model, the weight values in matrices such as Query, Key, and others generally follow Normal distribution, but their distribution shapes differ. Within model the distributions among different projection matrices vary. Between different model families, or between models with same family but different scales, the distributions of the same kind of matrix are also not the same. In order to quantify these comparisons, according to Normal distribution function for these weight data, the relevant parameters of function are calculated. Assuming the weight values $w_1$, $w_2$,....$w_i$,… that follow Normal distribution are as following:

$$W = \{w_1, w_2 ..., w_n\}, w_i \sim N(\mu, \sigma^2),$$

It is obvious that the value of μ is 0 in each model. The standard deviation values, which have significant impact on distribution, vary for each Normal distribution function. We calculate the standard deviation values of every matrix such as $std_Q$ for Query-Matrix, $std_{Down}$ for Down-Matrix and combine them into **Standard-Deviation Vector** denoted as

$$V_{std} = \{std_Q, std_K, std_V, std_O, std_{Gate}, std_{Up}, std_{Down}\}$$

The std values in the vector of a model have been normalized. From the results it is clear that both differences and connections among $V_{std}$ (Standard-Deviation Vectors) across different models. Specifically, the $V_{std}$ values differ significantly between models with distinct family. The models within the same family exhibit marked similarity in their $V_{std}$ values. This is illustrated in Figure 3. Figure 3(a) shows that in three models with similarly scale-SmolLM2-1.7B, LLaMA3.2-1B and Qwen2-1B, the $V_{std}$ vary considerably. SmolLM2 exhibits comparatively larger standard deviations across all projection matrices. In other two models, the std values peak at the Key-matrix and then decline. In Qwen2 the V, O, Gate, Up and Down matrices cluster at low standard deviation values levels. In LLaMA3 the standard deviation values at these points are relatively high. Data shows that the std of Key-Matrix in the Qwen2 differs significantly from std of other matrices, while the std of LLaMA3 are relatively close to each other. Figure 3(b) shows that the shapes of the curves formed by values in vectors are basically the same among LLaMA3-1B/3B/8B models. The $V_{std}$ curve shapes of the 1B/8B models are closer to each other, while the 3B model has some points which are higher. The curve chart indicates that the std values of each matrices are more different in 1B/8B models. The curve graphs of $V_{std}$ in different SmolLM2 and Qwen2 models are shown in Appendix Figure 13. The results are similar to $V_{std}$ in LLaMA3 models, with the same models family having similar shape of Standard-Deviation Vector.

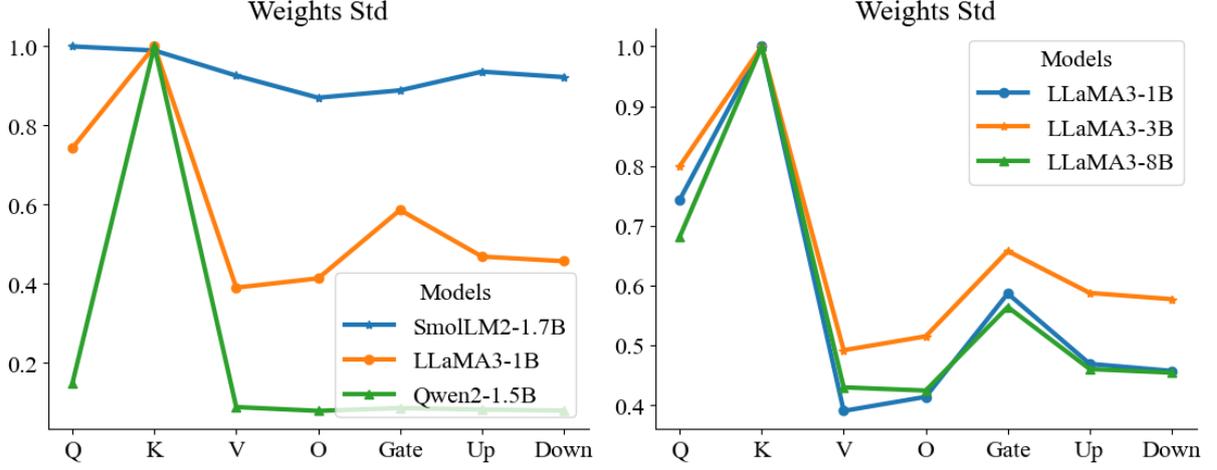

Figure3 (a) Standard-Deviation Vector of different models    (b) Standard-Deviation Vector of 1B/3B/8B LLaMA models

The preceding discussion investigates the characteristics based on weight–distribution relationships among different models and among the various projection matrices within model. The analysis shows that the weight values in models all exhibit the Normal distribution. However, the shapes of distribution differ across different matrices and models. The Standard-Deviation Vector reveals the structural characteristics of model's weights. These characteristics show similarity among models within the same family, while exhibiting disparities between different models.

## 3. Clustering of weight

### 3.1 Clustering Singular Values

Weight distributions and standard deviations characterize the properties of the whole model. To inspect structural signatures at finer granularity, we perform singular-value decomposition (SVD)(Klema et al., 1980) on every projection matrix. Taking LLaMA3.2-1B as an example, the 16 transformer layers each contain seven projection matrices (Q, K, V, O, Gate, Up, Down). After decomposing each matrix, we obtain its singular values. Because the matrices have different sizes, their singular values also differ in length. To enable comparison, we introduce a rank parameter and retain only the top-rank singular values as a concise feature of the matrix. Setting rank = 16 yields 16 × 7 = 112 singular-value vectors extracted from LLaMA3-1B. These vectors are collected into the $SET_{proj}$. The $S\_Q_i$ denoted as the singular-value vectors extracted from Query-proj matrix of the i-th layer in model.

$$SET_{proj} = \{S\_Q_1, S\_K_1, S\_V_1, \ldots, S\_V_{16}, S\_O_{16}, S\_Gate_{16}, S\_Up_{16}, S\_Down_{16}\}$$

We apply K-Means algorithm(Hartigan et al., 1976)(Virtanen et al., 2020) to cluster all vectors in $SET_{proj}$, repeating the procedure with randomized centroids for n_cluster = 2 and 3. Before clustering each singular value vector is normalized. To visualize the clustering result we first reduce the vectors to two dimensions with PCA. The scatter plot is shown in Figure 4. Comparing the two partitions, we select n_cluster = 2 for all subsequent analysis.

We label the classification of all vectors in $SET_{proj}$ with 0 and 1, as shown below.
[1 1 0 1 0 1 0 1 1 0 1 1 1 0 1 1 0 0 1 0 0 1 1 0 0 1 0 0 1 1 0 0 1 0 0 1 1 0 0 1 0 1 1 1 0 0 1 0 1 1 1 0 0 1 0 1 1 1 0
0 1 0 0 1 1 0 0 1 0 1 1 1 0 0 1 0 0 1 1 0 0 1 0 0 1 1 0 1 1 0 0 1 1 0 1 1 0 0 1 1 0 1 1 1 0 1 1 0 0 1 1 0]

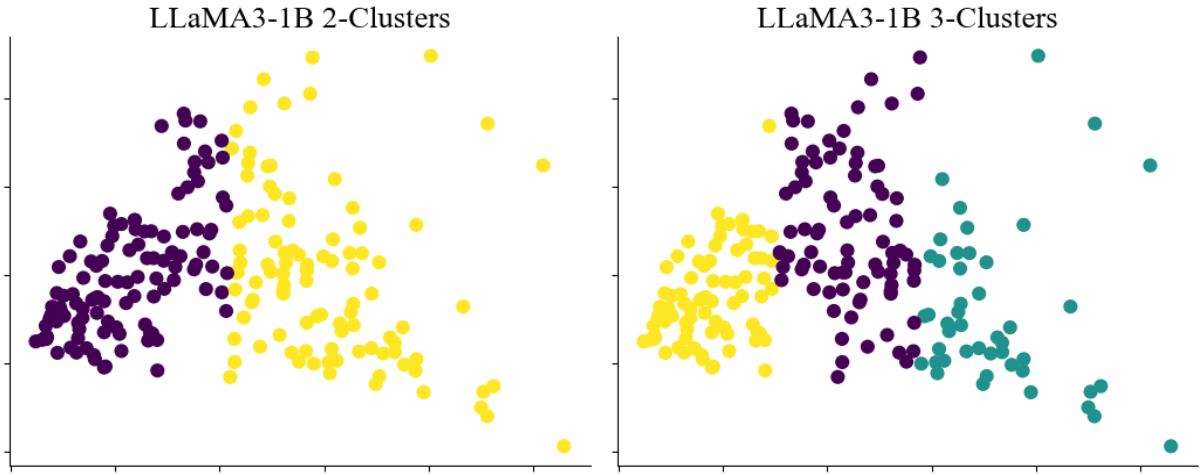

Figure4 (a)clustering LLaMA3-1B with 2 groups       (b) clustering LLaMA3-1B with 3 groups

The scatter plot of cluster labels is too rough to provide useful information. When we reorder the cluster assignments by layer index, however, a clear pattern emerges across projection types. Figure 5(a) displays the cluster labels for all 112 singular-value vectors in LLaMA3-1B, arranged layer-wise in the order of Q, K, V, O, Gate, Up and Down. It can be seen in the diagram that although clustering is performed for each vector, the results show that the projection matrices types corresponding to the vectors have distinct categories. Among them, all singular values of Q and K belong to the same category, while all singular values of V and most of the singular values of O, Up, and Down belong to another category. The clustering results are visualized as a heatmap in Figure 5(b). The light cells indicate label 1 and dark cells indicate label 0. Each row corresponds to one transformer layer; each column contains the category for every singular-value vector of a given projection matrix type (Q, K, V, O, Gate, Up, Down) across all layers. From the heatmap, the classification of singular values for each kind of projection matrix can be seen more clearly.

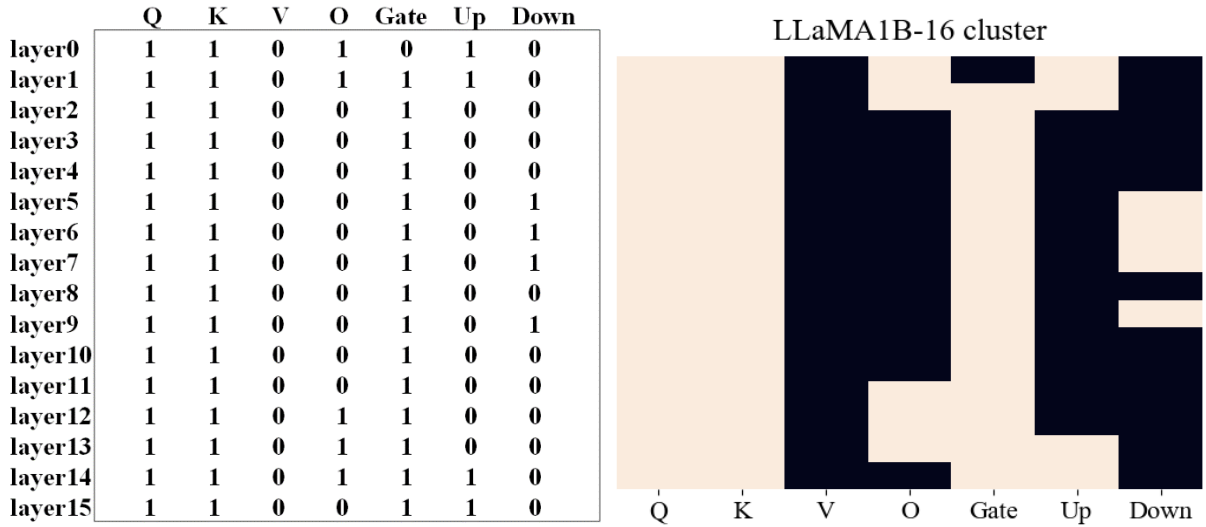

Figure5 (a)cluster labels of LLaMA3-1B       (b) heatmap of clustering LLaMA3-1B-16rank with 2 groups

The clustering described above is performed on singular-value vectors truncated to rank = 16. Repeating the experiment with rank = 64 yields the heatmap shown in Appendix Figure 14(a). The clustering results are basically the same as rank=16, with some individual vectors having changed categories. Heatmaps for LLaMA-3-8B and Qwen2-1.5 B/7 B are shown in Appendix Figure 14(a)(c)(d). The results show that although the singular value clustering results of different models are not the same, most singular values belonging to the same projection matrix have a clear distinction from each other.

## 3.2 Clustering Vector

For the above analysis, the clustering features of each model are represented in a quantitative manner. We use K-Means to classify the extracted singular values into two categories. Then the clustering labels are divided into seven groups according to the type of their projection matrix: Q, K, V, O, Gate, Up, and Down. The clustering labels of the same type as that in Query-matrix are recorded as 1, while the labels in another category are recorded as 0. The mean for the classification labels of all elements in each group is calculated, and the results are recorded as Clustering Vector. In LLaMA3-1B the clustering vector denoted as $CV_{\text{LLaMA}}$ as follow.

$$CV_{\text{LLaMA}} = \{ C_Q, C_K, C_V, C_O, C_{\text{Gate}}, C_{\text{Up}}, C_{\text{Down}} \}$$

$C_Q$ represents the mean of all the singular value clustering labels of the Query-Matrix in the model. The clustering vector curves of LLaMA3-1B/3B/8B models under rank=16 and rank=64 are shown in Figure 6. The results indicate that the clustering vector shapes are the same across different scales of all LLaMA3 models. The clustering vector curves of Qwen2 models and SmolLM2 models are shown in Appendix Figure 15. The results show that both Qwen2 and SmolLM2 models have their own distinct characteristic for clustering vectors.

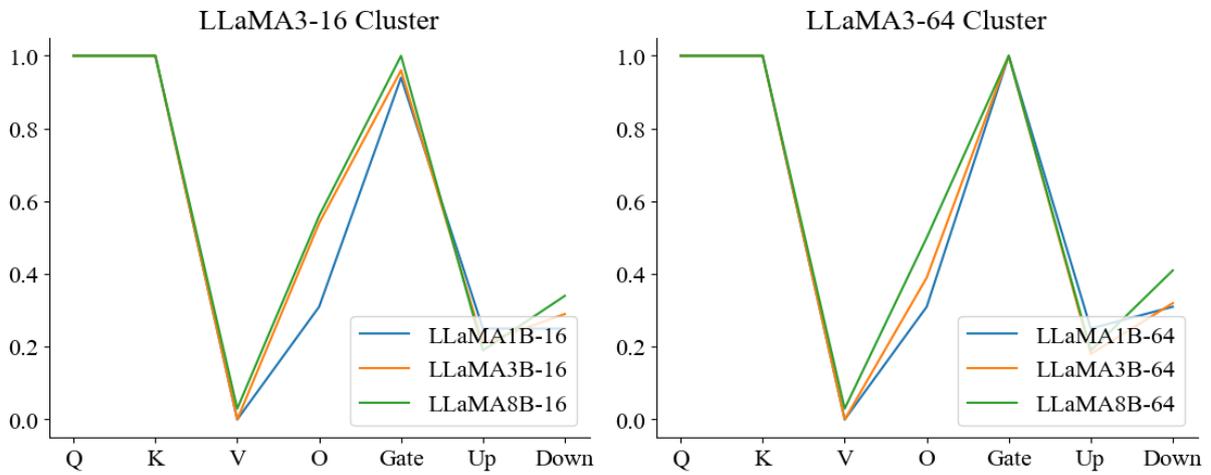

Figure6 (a)Clustering Vector of LLaMA3 with rank=16   (b) Clustering Vector of LLaMA3 with rank=64

The result now provides evidence that weight in LLMs approximately follow Normal distribution, but the parameters of distribution differ across model families. Within a single model the weight distributions also vary from one kind of projection to another. Standard deviation is the parameter that directly affects shape of the distribution. So, we proposed by computing the standard deviation of every projection matrix and arranging the values in the fixed order Q, K, V, O, Gate, Up, Down to form vectors, which are used to analyze characteristics of weight. We find these vectors are highly similar among models of the same family and clearly distinct across model families.

To probe more inter-weight relationships, we perform k-means clustering on the singular values of all projection matrices, aggregate the cluster labels by projection type, and average them to obtain the clustering vector. It is found that models from the same family yield nearly identical clustering vector, whereas different families produce markedly different vectors.

We believe that both vectors can demonstrate the characteristics of the model from different angles. When the weight values change, the impact on the standard deviation vector becomes more pronounced. This vector is more inclined to describe characteristic of model as a whole. As training reshapes the weights, the standard-deviation vector is liable to shift, reflecting the altered distributional spread. In contrast, the clustering vector encodes the relationships among Query, Key, etc. This vector is tied to the model's architecture or working principle. So it may remain comparatively stable during training.

## 4. Characteristics for LoRA

We analyze how the two vector characteristics shift after LoRA fine-tuning by direct experiment. The pre-trained models are LLaMA3.2-1B and SmolLM2-1.7B. The datasets for LoRA fine-tuning are alpaca-gpt4-data-en(Peng et al. 2023) and reasoning-1-1k[1]. Training parameters can be found in Appendix Table1. For the A and B matrices corresponding to the projection matrix in each layer after training, we calculate B*A to generate weight data with the same size as the original projection matrix. The standard deviation vector and clustering vector in the aforementioned manner are extracted from these weight data and are used to compare the results under different models and different dataset training. Two experiments are designed. (1) Fix the dataset and vary only the pre-trained model;(2) Fix the pre-trained model and vary only the dataset.

### 4.1. Same dataset for different models

We first examine how the same dataset influences LoRA training for different models. Figure 7 displays the clustering vectors of models after LoRA training on the reasoning-**1k** dataset and Alpaca-**Gpt** dataset. The results in the Figure show no obvious correlation between the vectors of each model. Figure 8(a) plots the standard deviation vectors of two models after LoRA training on reasoning-1k. It can be seen from the figure that the two curves are nearly identical. Figure 8(b) shows the same metric after training on the Alpaca-GPT dataset, and a similar conclusion is observed.

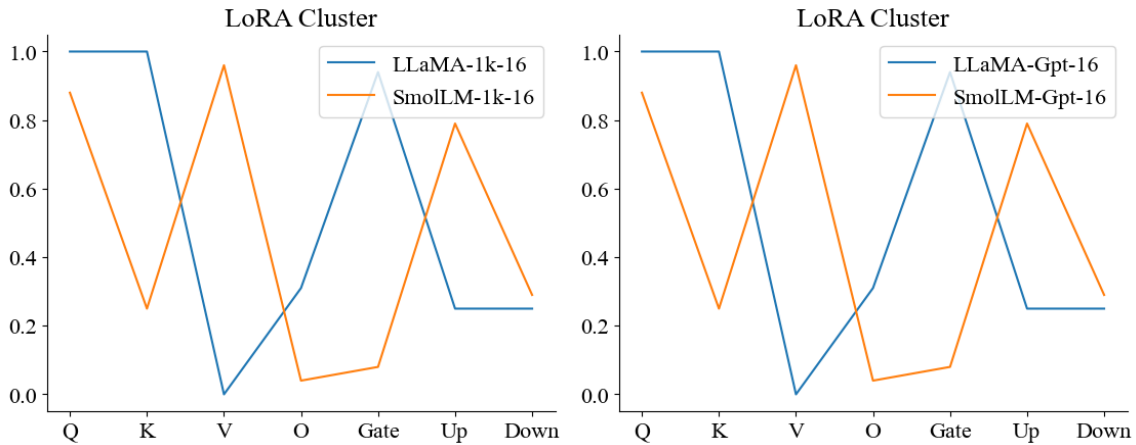

Figure7 (a) clustering vectors with dataset reasoning-1K    (b) clustering vectors with dataset Alpaca-Gpt

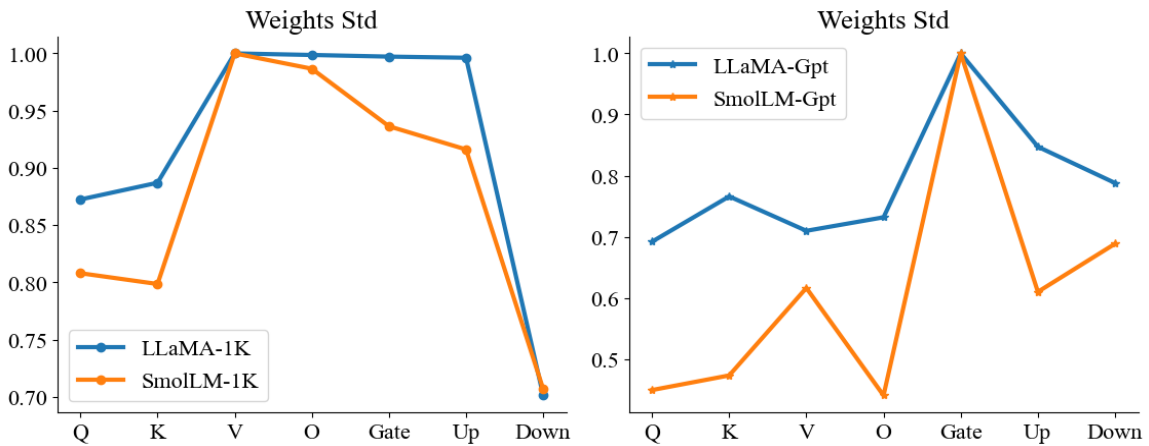

Figure8 (a) std vectors with dataset reasoning-1K    (b) cluster vector with dataset Alpaca-Gpt

The experimental results indicate that different pretrained models, when fine-tuned on the same dataset via LoRA, yield weight distributions that are strikingly similar. In the figures their standard deviation vectors closely overlap. This suggests that the distribution of LoRA-induced weights is governed primarily by the training dataset, independently of the original weight distribution of the pre-trained model. In contrast, the clustering vectors remain unaffected, implying that the correlations between different weight are left untouched by the training dataset.

---
[1] https://huggingface.co/datasets/fluently-sets/reasoning-1-1k

### 4.2. Different dataset with same model

We analyze the impact of LoRA training on weights using different datasets with the same model. Figure 9(a) compares the standard deviation vectors for original weights of the LLaMA3-1B model itself with those after LoRA training. The three curves are all distinct and show no obvious regularity. Figure 9(b) shows the comparison of the standard deviation vectors after training the SmolLM2-1.7B model, with results being the same as in Figure 9(a). Figure 10 displays the comparison of clustering vectors of weights after LoRA training on two models. The results indicate that the trained clustering vectors are completely identical as the original model's clustering vectors, hence the curves overlap. Heat-maps of the model weights are provided in Appendix Figure 16.

In the above experiment, the same dataset is used for fine-tuning on different models, and different datasets are used for fine-tuning on the same model. The results indicate that the distribution shape of weights after training is not related to the original model, but rather is related to the dataset used. However, the clustering vector, which is used to illustrate relationship between different weights, is the same as that of the original pre-trained model and is not influenced by the training data.

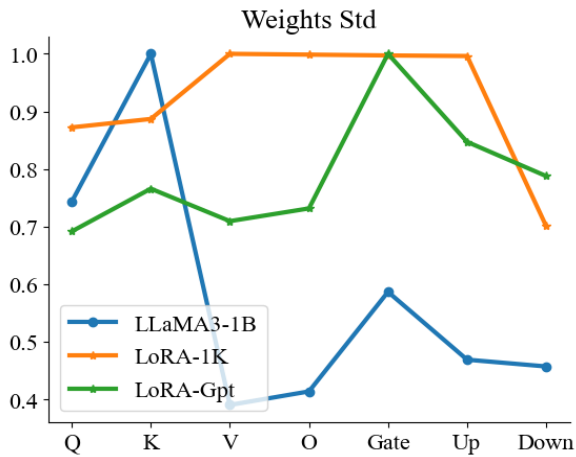

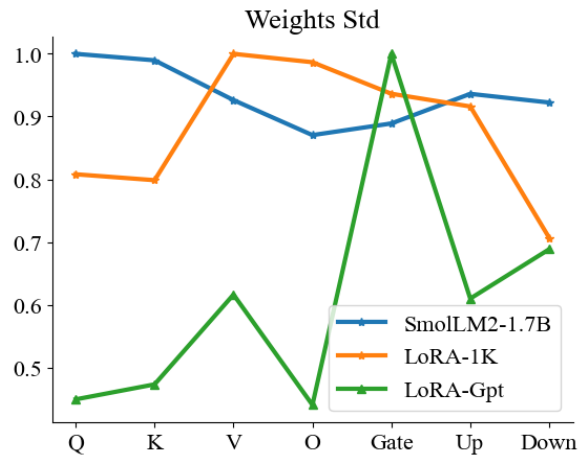

Figure9(a) std vectors of LLaMA-1B  (b) std vectors of SmolLM-1.7B

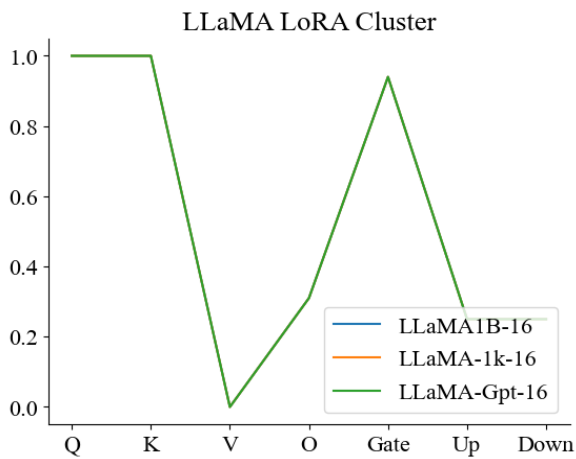

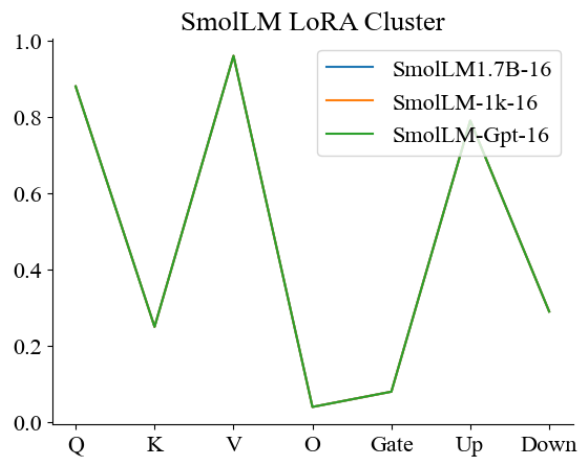

Figure10 (a)cluster vector of LLaMA3 rank=16  (b) cluster vector of LLaMA3 rank=64

# 5. Conclusion

This paper investigates the weight characteristics of large language models with the aim of exposing both intra- and inter-model relationships that can help for performance improvements. We distill two compact descriptors from the parameter tensors: Standard Deviation Vector that encodes the Normal distribution of weight in different projection matrices. Clustering vector that encodes how the singular values of those matrices group within models. These two vectors can serve as weight-level fingerprints to distinguish among different models. Within the same model family, both the standard deviation vector and the clustering vector exhibit high similarity, while across model families they diverge markedly. LoRA fine-tuning experiments further reveal that standard deviation vector rapidly reshapes to mirror the weights influenced by training dataset. Different models converged on the same dataset acquire the similar standard deviation vector. By contrast, the clustering vector remains frozen to the pre-trained model, implying that data perturbs variance but not inter-weight correlational structure. Future work will continue to analyze the role of these characteristics in model improvement.

# Appendix

## A. Experiments details

**Table 1:** Configuration for Experiments

| Argument | Setting |
| --- | --- |
| OS & device | Ubuntu 24.04.1 LTS<br>NVIDIA GeForce RTX 4070 Laptop GPU 8G |
| per device train batch size | 2 |
| gradient accumulation steps | 2 |
| num_train_epochs | 1 |
| dropout | 0.2 |
| learning rate | 1e-4 |
| warmup ratio | 10 |

## B. Figures of other models

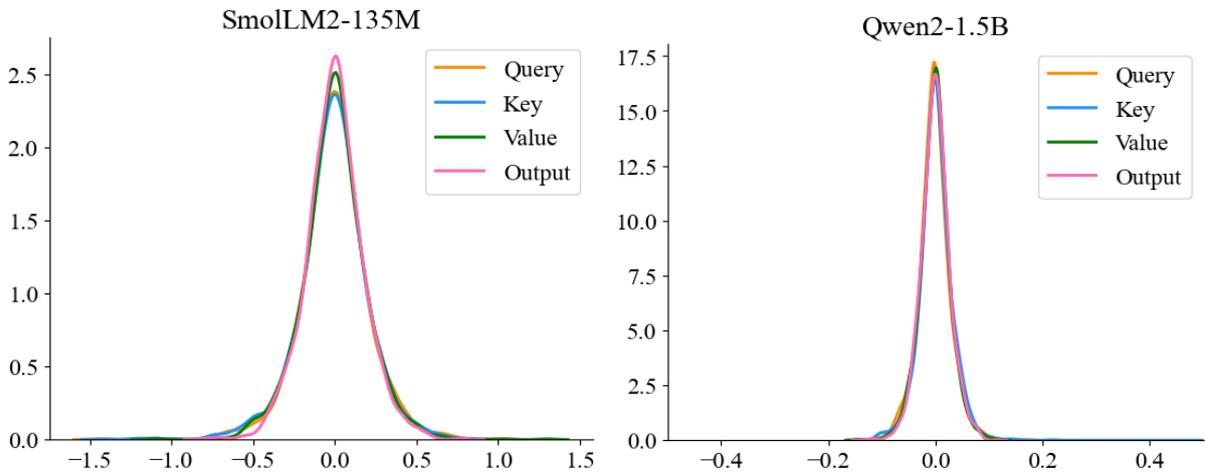

(a) Distribution of SmolLM2-135M(Q/k/V/O)   (b) Distribution of Qwen2-1.5B(Q/k/V/O)

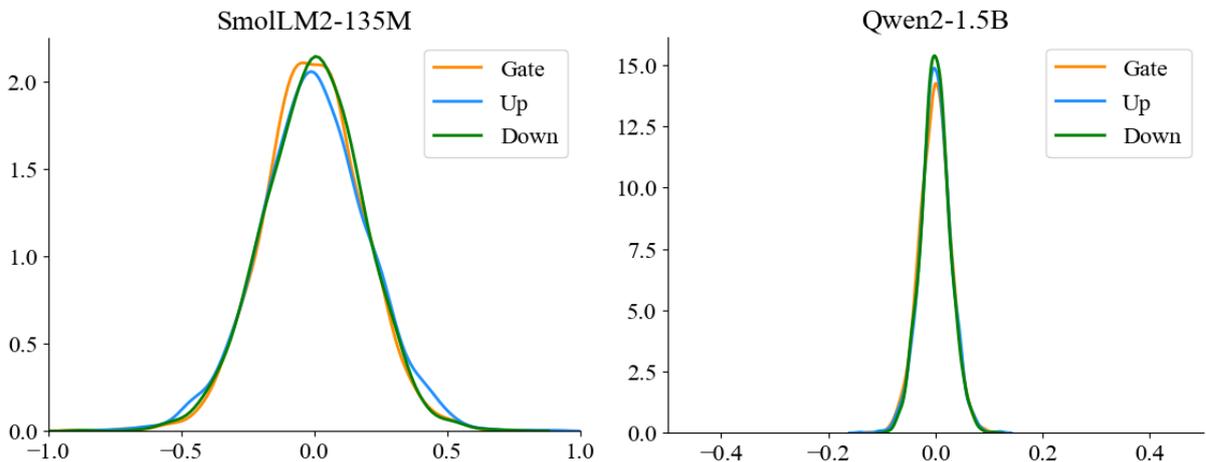

Figure11 (c) Distribution of SmolLM2-135M(gate/up/down) (d) Distribution of Qwen2-1.5B(gate/up/down)

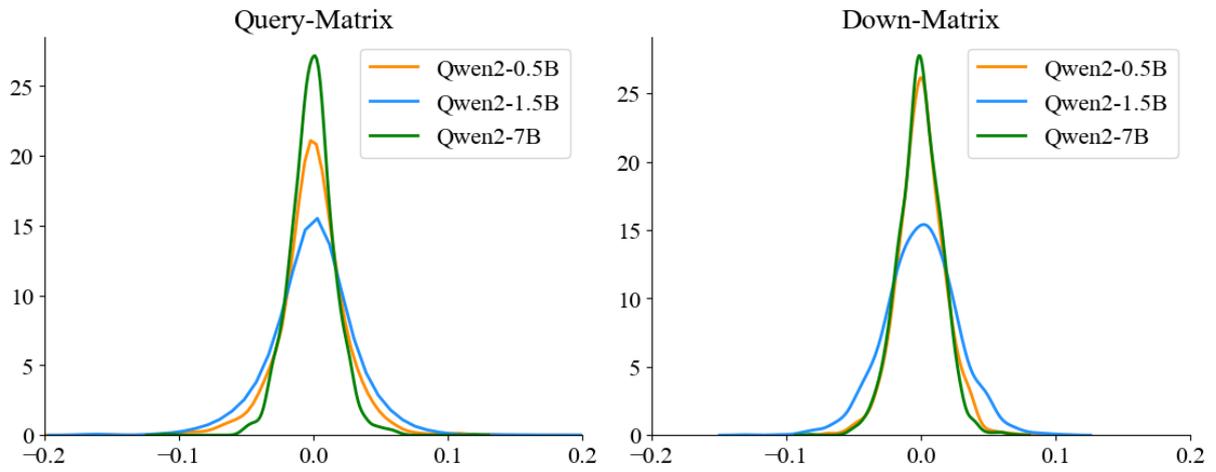

Figure12 Distribution for Qwen2 models

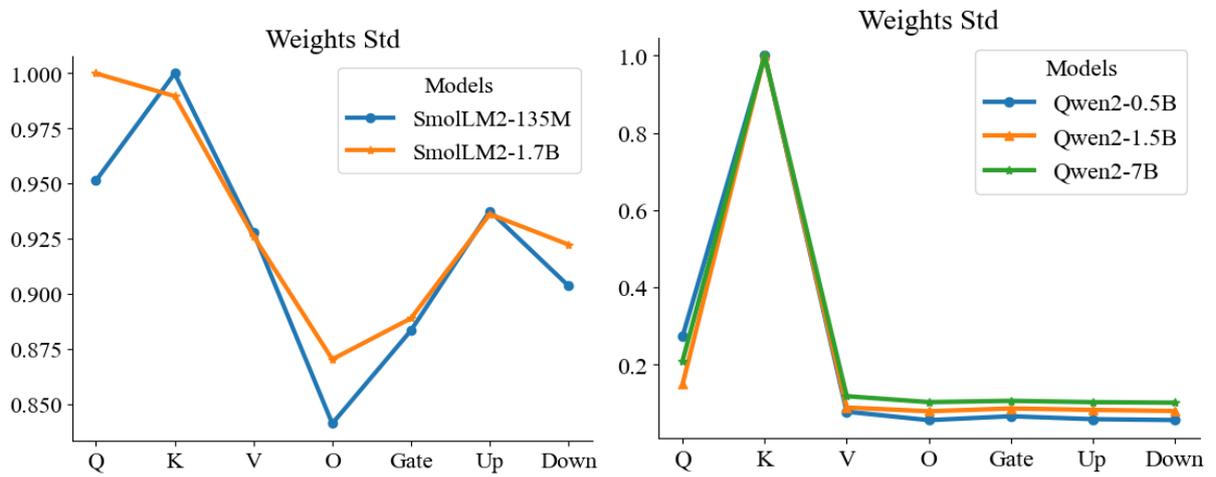

Figure13 (a) Normalized std of SmolLM models  (b) Normalized std of Qwen models

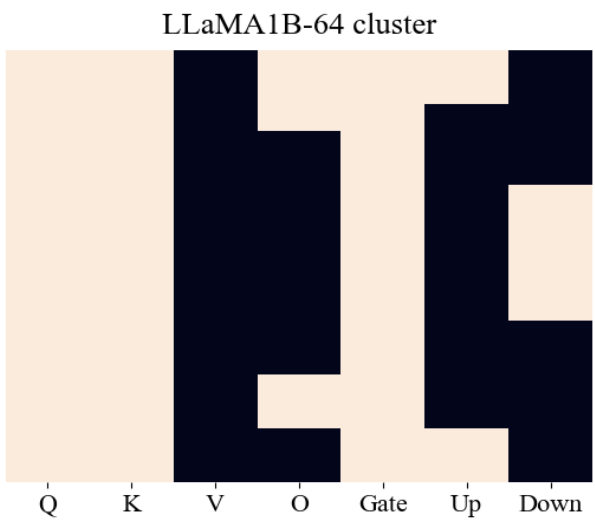
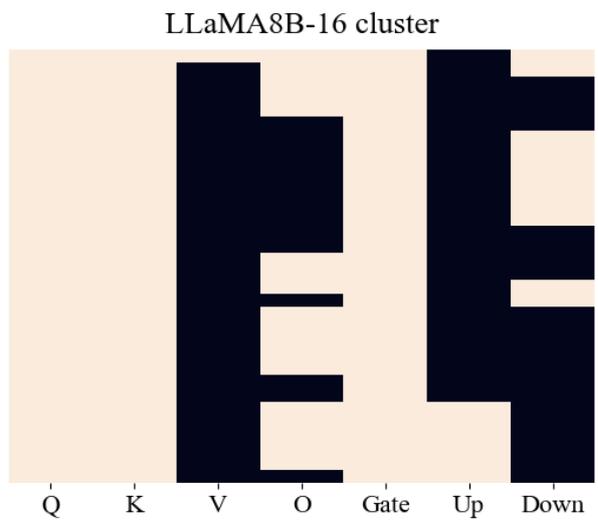

(a) heatmap of clustering LLaMA3-1B-64rank  (b) heatmap of clustering LLaMA3-8B-16rank

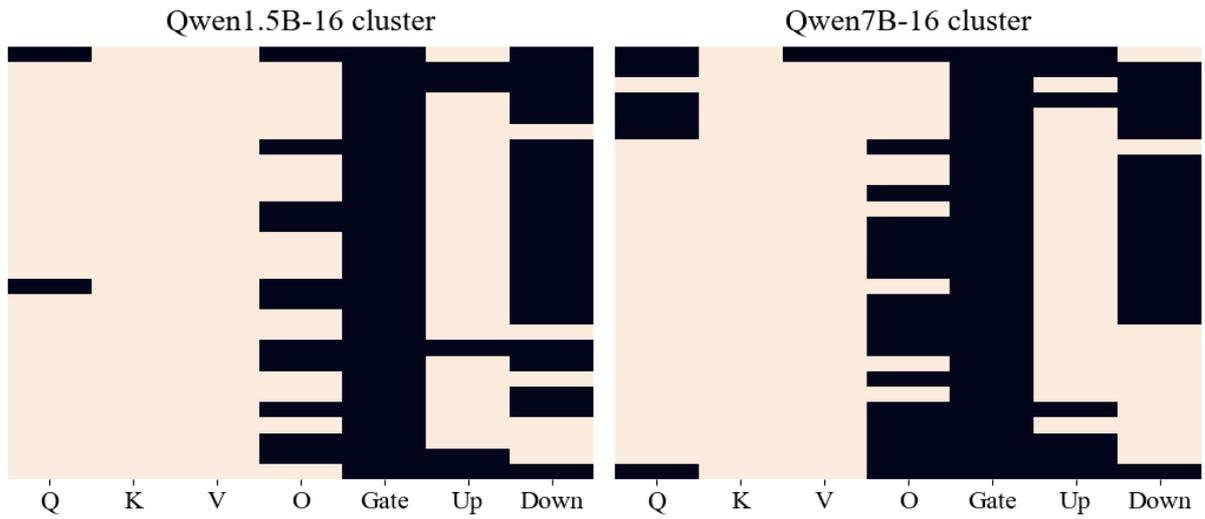

Figure14 (c) heatmap of clustering Qwen-1.5B    (d) heatmap of clustering Qwen-7B

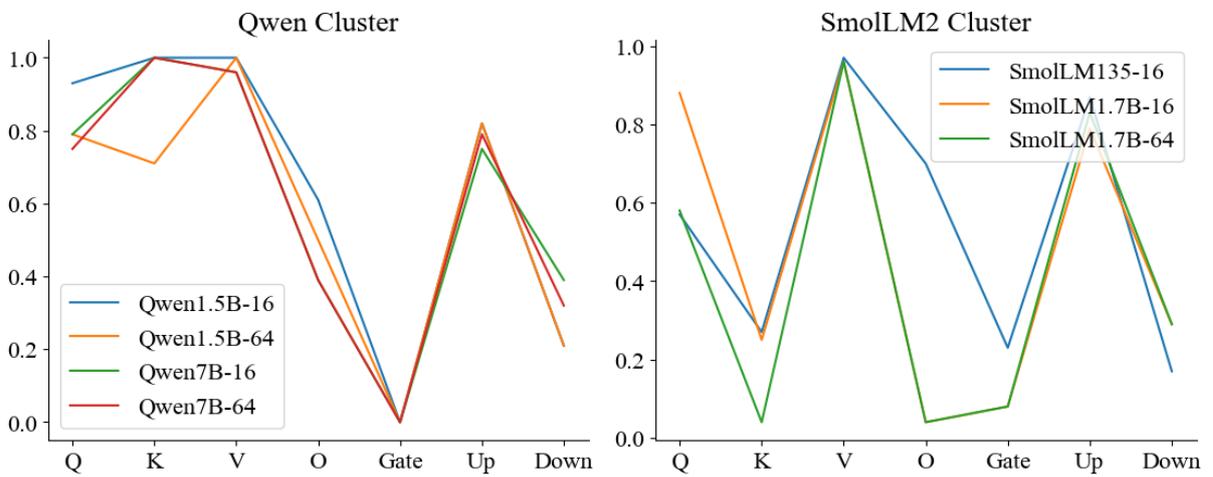

Figure15 (a) cluster vector of Qwen2 models    (b) cluster vector of SmolLM models

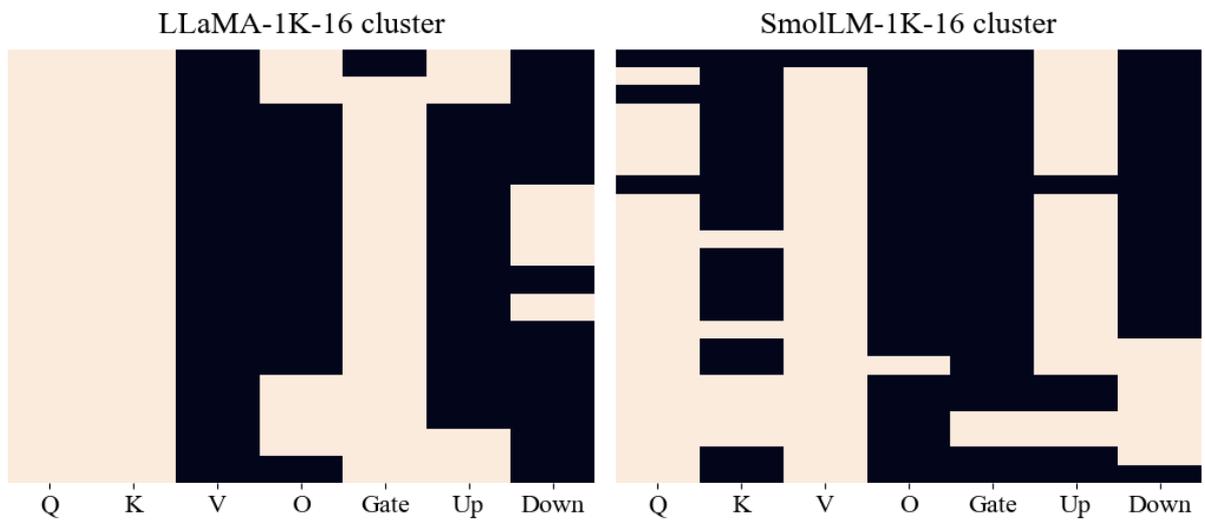

Figure16 (a) cluster vector of LLaMA3 rank=16    (b) cluster vector of LLaMA3 rank=64